\documentclass[letterpaper, 10 pt, journal, twoside]{IEEEtran}
\IEEEoverridecommandlockouts 
\usepackage{graphicx}
\usepackage{bbold}
\usepackage{algorithm}
\usepackage{algorithmicx}
\usepackage[noend]{algpseudocode}
\usepackage{cite}
\usepackage{color}
\usepackage[dvipsnames]{xcolor}
\usepackage{amsmath,amssymb,amsfonts,dsfont}
\usepackage{bm}
\usepackage{setspace}
\usepackage{multirow}
\usepackage{mathtools}
\usepackage{fancyhdr}
\usepackage{authblk}
\usepackage{url}
\usepackage{tikz}

\usepackage{hyperref}

\usepackage[
  separate-uncertainty = true,
  multi-part-units = repeat
]{siunitx}
\usepackage{threeparttable}
\usetikzlibrary{decorations.pathreplacing}

\usepackage{renew}      
\usepackage[switch]{lineno} 
\usepackage{amsmath}   
\newcommand{\loss}[1][{}]{{\mathcal{L}}_{\rm #1}}

\title{\LARGE \bf
SGL: Symbolic Goal Learning in a Hybrid, Modular Framework
for Human Instruction Following
}

\author{Ruinian Xu, Hongyi Chen, Yunzhi Lin, and Patricio A. Vela
\thanks{Ruinian Xu, Hongyi Chen, Yunzhi Lin and Patricio A. Vela are with Institute for Robotics and Intelligent Machines, 
    Georgia Institute of Technology, GA, USA.
    {\tt\small \{rxu72, hchen657, yunzhi.lin, pvela\}@gatech.edu}}%
\thanks{* This work was supported in part by NSF Award \#2026611.}%
}

\begin{document}

\maketitle
\thispagestyle{empty}
\pagestyle{empty}

\begin{abstract}
This paper investigates robot manipulation based on human instruction with
ambiguous requests.  
The intent is to compensate for imperfect natural language 
via visual observations.
Early symbolic methods, based on manually defined symbols,
built modular framework consist of semantic parsing and task
planning for producing sequences of actions from natural language requests.  
%
%
Modern connectionist methods employ deep neural networks to 
automatically learn visual and linguistic features and map to 
a sequence of low-level actions, in an end-to-end fashion. 
%
%
These two approaches are blended to create a hybrid, modular framework: 
it formulates instruction following as symbolic goal learning
via deep neural networks followed by task planning via symbolic
planners. 
Connectionist and symbolic modules are bridged with 
Planning Domain Definition Language.
The vision-and-language learning network predicts its goal representation,
which is sent to a planner for producing a task-completing action sequence. 
For improving the flexibility of natural language, we further 
incorporate implicit human intents with explicit human instructions.
To learn generic features for vision and language, we propose to separately
pretrain vision and language encoders on scene graph parsing and semantic
textual similarity tasks.
Benchmarking evaluates the impacts of different components of, or options
for, the vision-and-language learning model and shows the effectiveness of
pretraining strategies.
Manipulation experiments conducted in the simulator AI2THOR show 
the robustness of the framework to novel scenarios.
\end{abstract}


\section{Introduction}
\IEEEPARstart{I}{deally} robot agents sharing the same working space with humans and assisting
them would be capable of interpreting human instructions and performing
their corresponding tasks.
Human instruction following is a long-standing topic of interest whose
main challenge comes from the diversity of communication and interpetation,
which permits incomplete or ambiguous natural language. 
This paper proposes to disambiguate natural language via visual information
with a hybrid, modular framework.

Early symbolic works employ semantic parsing and task planning to 
first map natural language into certain representations and then generate
a sequence of actions. 
Attempts to address the ambiguity of natural lanugage include
incorporating knowledge bases \cite{tenorth2013knowrob, antunes2016human}, 
dialogue systems \cite{pramanick2019your}, 
and vision \cite{misra2016tell}.
Semantic parsing, which relies on syntactic structure of natural language,
can't well capture its semantic meaning and has the issue with abstract or vague
language input such as human intents. 
%

The rise of connectionist approaches provided a means to avoid processing
natural language and vision based engineered symbolic representations, by 
automatically learning visual and linguistic features via deep 
neural networks.
For example, sequence-to-sequence models learn to map raw vision and
language input into a sequence of low-level actions
\cite{nazarczuk2020v2a, shridhar2020alfred}. 
These end-to-end designs suffer from performance drops in testing stage.
To avoid all-in-one network designs, researchers have sought to factorize
the network into sub-modules or separately consider different types of
tasks \cite{dipendra2018mapping, singh2021factorizing,corona2020modular,
zhou2021hierarchical}.

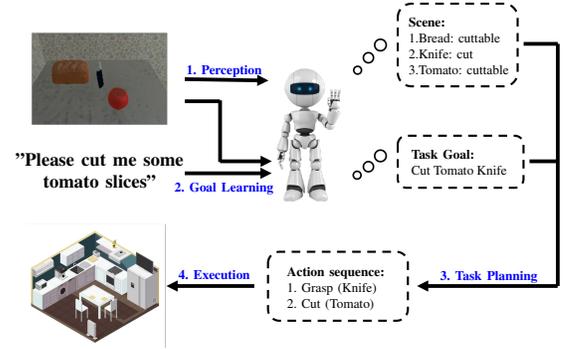
\begin{figure}[t]
  \centering
  \begin{tikzpicture} [outer sep=0pt, inner sep=0pt]
  \node[anchor=north west] (a) at (0in,0in) 
    {\includegraphics[width=0.8\columnwidth, clip=true,trim=0in 0in 0in 0in]{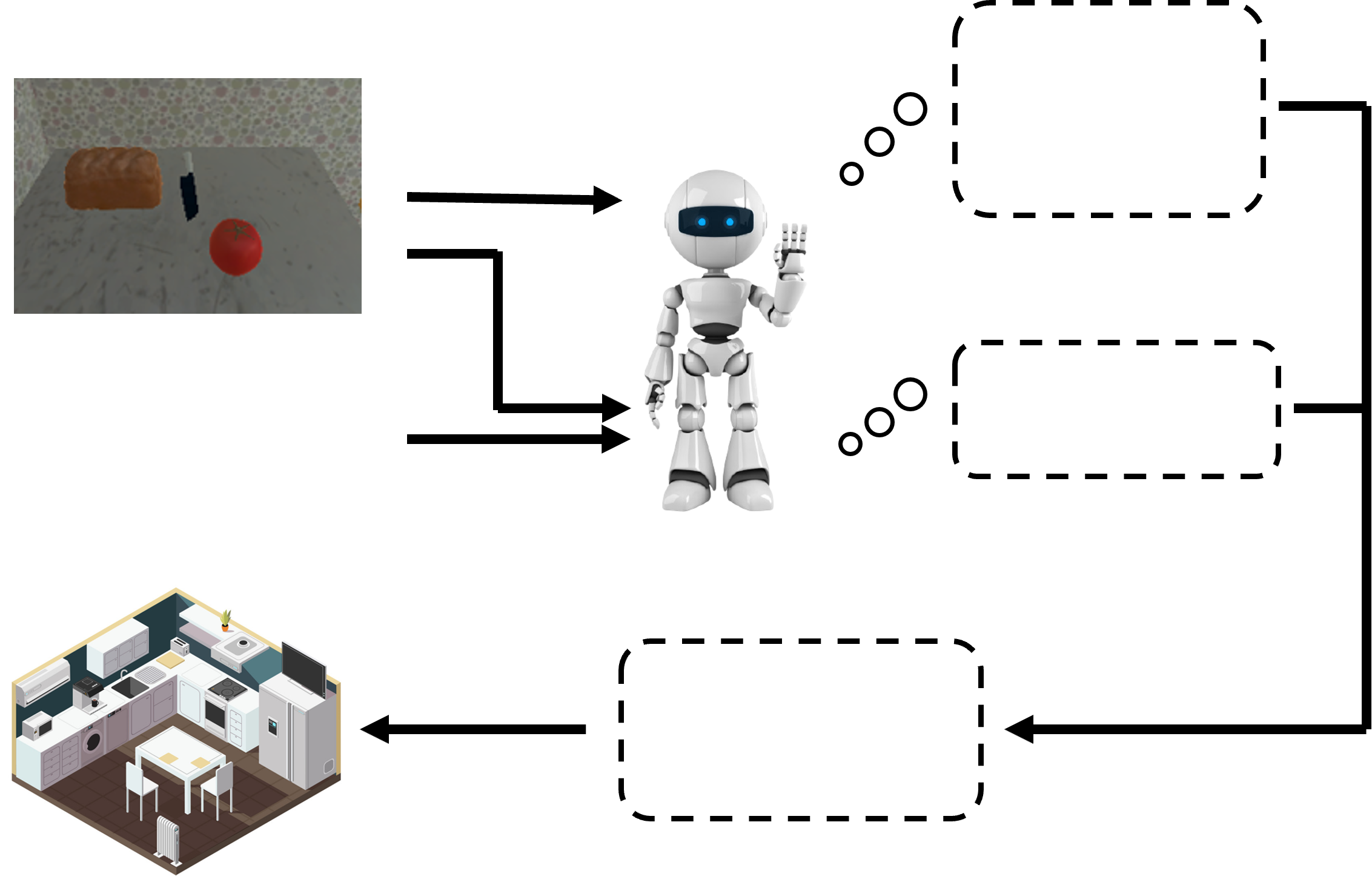}}; 
	\node[anchor=north west,xshift=57.5pt,yshift=-20.5pt,text=blue,inner sep=3pt] at (a.north west) {\tiny \bf 1. Perception};
	\node[anchor=north west,xshift=35.5pt,yshift=-65.0pt,text=blue,inner sep=3pt] at (a.north west) {\parbox{1.0in}{\tiny \bf \centering 2. Goal Learning}};
	\node[anchor=north west,xshift=153.5pt,yshift=-98.5pt,text=blue,inner sep=3pt] at (a.north west) {\tiny \bf 3. Task Planning};
	\node[anchor=north west,xshift=54.5pt,yshift=-98.0pt,text=blue,inner sep=3pt] at (a.north west) {\tiny \bf 4. Execution};
    
	\node[anchor=north west,xshift=-11.5pt,yshift=-54.0pt,text=black,inner sep=3pt] at (a.north west) {\parbox{1.0in}{\scriptsize \bf
    \centering "Please cut me some tomato slices"}};
    
    \node[anchor=north west,xshift=141.5pt,yshift=-2.5pt,text=black,inner sep=3pt] at (a.north west) {\parbox{0.7in}{\tiny \textbf{Scene:} \\1.Bread: cuttable \\2.Knife: cut \\3.Tomato: cuttable}};
    \node[anchor=north west,xshift=142.5pt,yshift=-52.5pt,text=black,inner sep=3pt] at (a.north west) {\parbox{0.5in}{\tiny \textbf{Task Goal:} \\ Cut Tomato Knife}};
    \node[anchor=north west,xshift=95.5pt,yshift=-97.0pt,text=black,inner sep=3pt] at (a.north west) {\parbox{1.0in}{\tiny \textbf{Action sequence:} \\ 
    1. Grasp (Knife) \\
    2. Cut (Tomato)}};

  \end{tikzpicture}
  \caption{Illustration of a hybrid and modular framework for 
  human instruction following. 
  The framework consists of four main components which are perception,
  goal learning, task planning and execution.  
  Better view in color.
  \label{fig_overview}}
  \vspace*{-1.25em}
\end{figure}

To leverage strengths of symbolic and connectionist approaches to compensate
limitation of each other, we propose a hybrid, modular framework
as depicted in Figure \ref{fig_overview}. 
It consists of perception, goal learning via a deep network, task planning 
via a task planner and execution.
Inspired by previous methods for manipulation task completion via object
affordance recognition \cite{chu2019toward, chu2020recognizing}, 
we formulate goal learning as predicting symbolic goal representation
for Planning Domain Definition Language (PDDL),
which bridges connectionist goal learning and symbolic task planning in
the proposed hybrid system.
In addition to incomplete human instructions, we further consider implicit
human intent, which has not been explored in existing methods. 
The learning phase of proposed visuo-linguistic network should benefit
from pretraining \cite{sun2019videobert,lu2019vilbert,li2019visualbert}
if task-relevant tasks can be identified.  Considering robotic
manipulation requires interpreting the interaction between objects, we
propose to pretrain the visual encoder on scene graph parsing. 
Though valid natural language requests can be structured quite
differently, such as explicit human instruction versus implicit human
intent, they should have similar semantic meaning.
Thus, the linguistic encoder is pretrained on a semantic textual
similarity task. 
This paper's main contributions are: \\
\textbf{(1)} A hybrid, modular framework for human instruction following.
The hybrid approach leverages the semantic feature learning properties
of deep neural networks and the symbolic computation of task planners.
The modular structure enables easy component-level analysis and 
upgrades. \\
\textbf{(2)} Benchmarking evaluates the impacts of different components for
vision-and-language learning.  The proposed strategy of separately
pretraining visual and linguistic encoders, on scene graph parsing and
semantic textual similarity tasks, outperforms standard pretraining
strategies. 
\\
\textbf{(3)} Manipulation experiments conducted in the simulator 
AI2THOR \cite{ai2thor}, with five daily activities and unseen scenarios, 
demonstrate the robustness of the proposed framework to novel objects and
environments.

\section{Related Work}


\subsection{Human Instrution Following}
Human instruction following requires robotic agents understanding 
human instructions and performing corresponding tasks. 
Common robotic task can be categorized into navigation and manipulation. 
%
%
Navigation and manipulation are tasks with different natures
in scene understanding.
Navigation focuses on identifying landmarks to understand where it is 
and where to go.
Manipulation requires interpreting interactions between objects and 
how to manipulate them.
Considering that, we will only study instruction following for 
robotic manipulation in this work. 
These review papers \cite{mogadala2021trends, tellex2020robots} 
well describe existing studies about Vision-and-Language Navigation. 
This section will first review existing symbolic and connectionist methods 
for human instruction following.
For connectionist method, a review is then provided for 
feature learning in vision-and-language 
joint learning deep neural networks. 
Common pretraining tasks are reviewed for learning
generic visual and linguistic features.

\subsubsection{Symbolic Method}
%
Human instruction following requires translating
human language into robot understandable language.
Based on manually defined symbols, early works employ semantic parsing 
to transform natural language into logical representations 
which perserves the meaning.
With well-structured input language, there are works parse natural language 
into formal semantic expressions such as a list of templates 
\cite{guadarrama2013grounding}, which is unscalable with 
the growth of the complexity of the manipulation task.
Instead of parsing natural language into formal representation, 
researchers have explored the direction of intermediate representations
such as Spatial Description Clause (SDC) \cite{tellex2011understanding}
and Linear Temporal Logic (LTL) \cite{kress2007structured, finucane2010ltlmop},
which will also be the direction of symbolic representation in this work.
%
%
However, instructions provided by non-expert human users can be vague 
or incomplete.
Realizing the ambiguity of natural language, 
some researchers attempt to incorporate external information 
such as knowledge base \cite{tenorth2013knowrob, antunes2016human}, 
dialogue system \cite{pramanick2019your}, 
visual information of surrounding environments \cite{misra2016tell} or
multi-source information \cite{lindes2017grounding}.
%
%
Among these auxiliary information, robotic vision, 
served as the simple and straightforward but rich way to 
disambiguate natural language, will be studied in this work.
%
%
Semantic parsing, which relies on syntax of language to perform
symbolic computation, can't well capture the semantic meaning of language 
and has the difficulty of translating abstract sentences such as 
human intents. 
Meanwhile, symbolic approaches using rule-based task planning achieve
high accuracy for computing action sequences for manipulation when the
symbols are correct.
%

\subsubsection{Connectionist Method}
With the significant evolution of connectionist methods in recent years,
deep neural networks show impressive strengths of learning semantic and 
high-dimensional features, which improves robustness to various types
of input data.
Packing everthing into one network, end-to-end learning models 
\cite{nazarczuk2020v2a, shridhar2020alfred}
are first proposed to directly map natural language and vision to
a sequence of low-level actions. 
%
%
The sequence-to-sequence model suffers from the well-known issue of 
teacher forcing, which leads to the poor performance under test scenarios. 
%
%
%
Observing the great performance drop from training to testing stages
of end-to-end learning models, 
researchers start to break the end-to-end network design and 
modularize the framework into several networks.
There are different designs of modular networks focus on different
natures of robotic tasks, such as factorizing the model into 
perception and action policy streams 
\cite{dipendra2018mapping, singh2021factorizing},
modularizing the model into separate sub-modules for sub-tasks
\cite{corona2020modular, zhou2021hierarchical},
decomposing the problem into sub-goal planning, 
scene navigation, and object manipulation \cite{zhang2021hierarchical}
and constructing the model into observation model, high-level controller 
and low-level controller \cite{blukis2022persistent}.
Modular networks significantly outperform end-to-end ones but the overall
performance in the testing stage requires further improvement.
The potential reason could be deep networks can't well maintain past 
information during the process of predicting sequential low-level actions.
%
%
%
%
%
%
%
Inspired by advantages and disadvantages of symbolic and connectionist 
apporoaches, we propose to address human instruction following via
a hybrid system.
%
%
Additionally, ambiguious natural language for complex manipulation tasks 
has not been explored in connectionist approaches, which will be another
aim in this work.

\subsection{Vision-and-Language Feature Learning}
Learning symbolic goal representation via vision-and-language deep
networks requires learning generic visual and linguistic features to
assist generalization to unseen scenarios. 
In this section, we provide a brief review of existing methods in
visual question answering for encoding visual and linguistic features
and corresponding pretraining tasks.

Visual feature learning method used in V\&L models can be categorized into 
Object Detector(OD)-based region, CNN-based grid and Vision Transformer(ViT) 
patch features. 
Due to the computational and time cost of pretraining vision transformer, 
this type of methods will not be explored and benchmarked. 
Most previous works \cite{anderson2018bottom, yu2019deep, 
li2019visualbert, lu2019vilbert} 
employ OD-based region features which are extracted via 
pretrained Faster R-CNN \cite{ren2015faster} based object detectors. 
The main concerns of this type of methods are frozen parameters and 
time cost of object detectors during the training and inference stage, 
respectively. 
To address above two issues, there are works 
\cite{jiang2020defense, huang2020pixel} 
have explored the way of extracting grid visual features via CNN 
such as ResNet \cite{he2016deep}, which makes the vision-and-language model 
end-to-end trainable. 
One-stage design for visual feature learning also reduces inference time but 
sacrifices a small amount of performance. 
%
%
For pretraining visual encoders to learn generic features, 
\cite{donahue2014decaf} found out that pretraining CNNs can be served as 
generic feature representation for many downstream tasks, 
such as object detection \cite{girshick2014rich}, 
semantic segmentation \cite{long2015fully}, 
and instance segmetnation \cite{hariharan2014simultaneous}.
Though existing pretraining tasks help capture object information
in imagery, they ignore potential interactions between objects
important to robotic tasks.

For linguistic feature learning, early research \cite{mikolov2013efficient, 
pennington2014glove, kiros2015skip} works on learning word-level feature
embeddings.
To learn high-level semantic embedding for sentences, based on
Recurrent Neural Networks (RNN), 
LSTM, Bidirectional LSTM, GRU \cite{Kyunghyun2014learning} and 
other similar designs are proposed.
The main concern of RNN-based methods is forgetting past information
for modeling long sequence data.
With the rise of Tansformers \cite{vaswani2017attention}, a series of 
approaches are proposed such as GPT \cite{radford2018improving}, 
BERT \cite{devlin2018bert}, RoBERTa \cite{liu2019roberta} and etc.
Among them, BERT model and its pretraining strategy of 
masked language modeling (MLM) is perhaps most widely used 
due to its simple network design and superior performance.
Modeling natural language without clustering sentences with 
similar semantic meanings, linguistic encoders might have 
the difficulty of interpreting similarity between 
explicit human instruction and implicit human intent.

The above literature review of symbolic and connectionist approaches
for human instruction following gives us the insight of 
a hybrid, modular system.
To leverage strengths of both methods and compensate each other's limitation, 
we propose to address human instruction following via connectionist
goal learning and symbolic task planning. 
Employing Planning Domain Definition Language (PDDL) as 
symbolic representation, we propose to bridge connectionist 
and symbolic approaches with symbolic goal representation of PDDL.
The vision-and-language connectionist framework, 
which consists of the visual encoder, linguistic encoder, multi-modal fusion 
and classification, is propose to learn symbolic goal representation.
The detected goal representation will be fed into symbolic task
planners to generate a sequence of actions.
To help learn generic features in vision-and-language framework,
we propose to separately pretrain the visual and linguistic encoders on
scene graph parsing and semantic textual similarity tasks.
Scene graph parsing forces visual encoders to capture relationships 
between objects, while semantic textual similarity helps linguistic encoders
learn similar semantic embeddings between human instructions and intents.
The modular design of goal learning and instruction following frameworks
enables simple replacement and upgrade for individual components
and also analysis for failures.

\section{Preliminaries}
\subsection{Planning Domain Definition Language}
For task planning, we employ the Planning Domain Definition Language (PDDL), 
a widely used symbolic planning language. 
With a list of pre-defined \textbf{objects} and their corresponding 
\textbf{predicates} (such as dirty, graspable, etc.),
a \textbf{domain} consists of primitive actions and corresponding effects.
Here, affordances and attributes serve to define available
\textbf{predicates} for subsequently specifying object-action-object 
relationships. 
Planning requires establishing a \textbf{problem}, which is composed of the 
initial state and a desired goal state of the world.
The initial state is formed with a list of objects with corresponding
predicates.
The goal state is structured in the form of action, subject and object.
From the \textbf{domain} and \textbf{problem} specification, a PDDL planner
produces a sequence of primitive actions leaving the world in the goal state
when executed.

\subsection{Problem Statement}
Given a RGB image and a sentence of natural language, the objective of this
framework is to generate a sequence of manipulation actions that achieve the
task indicated by the sentence.
Processing of RGB image generates an initial state estimate using an object
detector.
Completing the problem specification involves the proposed
vision-and-language deep learning framework, whose function is to convert
the paired image and natural language input into a symbolic goal
representation compatible with PDDL. 
Once the problem specification is built, the symbolic PDDL planner solves it to
generate the action sequence. The robot then performs the ordered actions in
the environment.

\section{Approach}
This section first describes the vision-and-language deep learning framwork 
proposed for learning symbolic goal representation for PDDL.
The framework is built in modular design.
Different approaches will be introduced for each module.
Two pretraining tasks are then proposed for learning generic visual and
linguistic features.
Lastly, a hybrid, modular framework for human instruction following 
is proposed to take vision and language as input and then output a sequence of 
actions to interact with environments.

\subsection{Vision-and-Language Task Goal Learning}
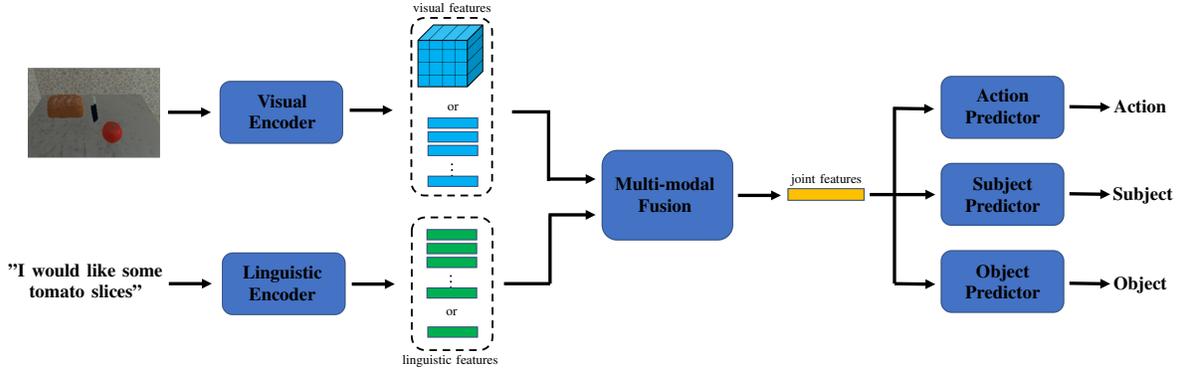
\begin{figure*}[t]
  \centering
  \begin{tikzpicture} [outer sep=0pt, inner sep=0pt]
  \node[anchor=north west] (a) at (0in,0in) 
    {\includegraphics[width=0.8\textwidth, clip=true,trim=0in 0in 0in 0in]{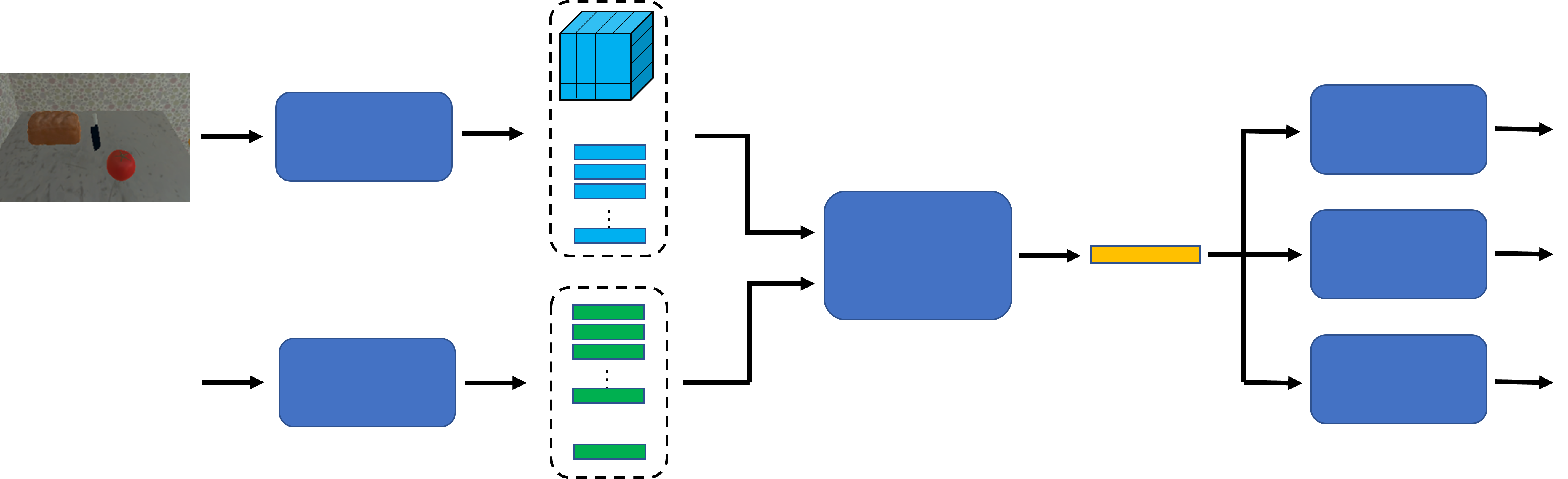}}; 
	\node[anchor=north west,xshift=-17.5pt,yshift=-90.0pt,text=black,inner sep=3pt] at (a.north west) {\parbox{1.0in}{\scriptsize \bf \centering "I would like some tomato slices"}};
	\node[anchor=north west,xshift=75.pt,yshift=-26.pt,text=black,inner sep=3pt] at (a.north west) {\parbox{0.5in}{\scriptsize \bf
    \centering Visual Encoder}};
    \node[anchor=north west,xshift=75.pt,yshift=-91.pt,text=black,inner sep=3pt] at (a.north west) {\parbox{0.5in}{\scriptsize \bf
    \centering Linguistic Encoder}};
    \node[anchor=north west,xshift=212.5pt,yshift=-57.5pt,text=black,inner sep=3pt] at (a.north west) {\parbox{0.7in}{\scriptsize \bf
    \centering Multi-modal Fusion}};
    
    \node[anchor=north west,xshift=347.5pt,yshift=-24.pt,text=black,inner sep=3pt] at (a.north west) {\parbox{0.5in}{\scriptsize \bf
    \centering Action Predictor}};
    \node[anchor=north west,xshift=347.5pt,yshift=-57.5pt,text=black,inner sep=3pt] at (a.north west) {\parbox{0.5in}{\scriptsize \bf
    \centering Subject Predictor}};
    \node[anchor=north west,xshift=347.5pt,yshift=-90.5pt,text=black,inner sep=3pt] at (a.north west) {\parbox{0.5in}{\scriptsize \bf
    \centering Object Predictor}};
    
    \node[anchor=north west,xshift=399.5pt,yshift=-28.0pt,text=black,inner sep=3pt] at (a.north west) {\parbox{0.5in}{\scriptsize \bf
    \centering Action}};
    \node[anchor=north west,xshift=400.5pt,yshift=-61.0pt,text=black,inner sep=3pt] at (a.north west) {\parbox{0.5in}{\scriptsize \bf
    \centering Subject}};
    \node[anchor=north west,xshift=399.5pt,yshift=-95.0pt,text=black,inner sep=3pt] at (a.north west) {\parbox{0.5in}{\scriptsize \bf
    \centering Object}};
	
	\node[anchor=north west,xshift=142.5pt,yshift=8.5pt,text=black,inner sep=3pt] at (a.north west) {\tiny visual features};
	\node[anchor=north west,xshift=138.5pt,yshift=-124.5pt,text=black,inner sep=3pt] at (a.north west) {\tiny linguistic features};
	\node[anchor=north west,xshift=285.5pt,yshift=-56.0pt,text=black,inner sep=3pt] at (a.north west) {\tiny joint features};
	
	\node[anchor=north west,xshift=155.5pt,yshift=-30.0pt,text=black,inner sep=3pt] at (a.north west) {\tiny or};
	\node[anchor=north west,xshift=155.0pt,yshift=-107.5pt,text=black,inner sep=3pt] at (a.north west) {\tiny or};

  \end{tikzpicture}
  \vspace*{-0.25em}
  \caption{Vision and language symbolic goal learning network architecture. 
   From RGB image and natural lanugage inputs, it outputs a PDDL goal
   state (action, subject and object).  Dark blue blocks represent
   components.  Light blue, green and yellow blocks represent visual,
   linguistic and joint features.  \label{fig_vltl}}
  \vspace*{-0.25em}
\end{figure*}

Given a RGB image $I$ and natural language string $L$, the vision-and-language
deep learning framework outputs a simple PDDL goal consisting of action $a$,
subject $s$ and object $o$.
The proposed PDDL symbolic goal learning framework, depicted in Figure
\ref{fig_vltl}, adopts a modular design and consists of a visual encoder,
a linguistic encoder, multi-modal fusion and classification modules. 
%
The visual and linguistic encoders are reponsible for learning
visual and linguistic features, respectively. 
Visual and linguistic features are embedded in different domains, which 
requires to be fused into joint features.
Lastly, joint features are fed to the classification module for predicting
the goal representation.
The classification module is a 2-layer Multi-Layer Perception (MLP)
with 256 hidden dimensions; it will not be covered. 

\textbf{Visual Encoder.} The visual encoder produces a set of local features
$V = \{v_1, \cdots, v_n\}$, from a RGB image $I \in \mathbb{R}^{H \times W
\times 3}$. 
There are two principal types of visual features, grid and region.
To generate the grid feature, the 2D image will be fed into a backbone
network, such as ResNet \cite{he2016deep}.
We treat each grid or pixel over the produced feature map 
$D \in \mathbb{D}^{H_1 \times W_1 \times C_1}$ as a local feature $v_i$,
where $H_1$ and $W_1$ are the height and width and $C_1$ is the dimension of
each feature.
To better localize the latent region, based on two-stage object detectors 
such as Faster R-CNN \cite{ren2015faster}, region features 
$V \in \mathbb{R}^{N \times C_1}$ 
are extracted from regions proposed by region proposal network,
where $N$ is the number of regions.

\textbf{Linguistic Encoder.} 
Given a natural language sentenc $L$ composed of $K$ words, the linguistic
encoder can either generate the corresponding embedding set $Q = \{q_1,
\cdots, q_k\}$ which represents each word or a single embedding vector $q$
which represents the semantic meaning of the entire sentence.
The encoder commonly involves word embedding and feature
encoding.
For word embedding, each word in the sentence will be mapped to an
embedding based on some pretrained embedding tables, such as
GloVe \cite{pennington2014glove}.
For feature encoding, LSTM, capable of connecting past information 
to the current task, has become a ubiquitous network model 
for language modeling.
However, due to issues arising from their sequential design and forgetting
past information, attention-based transformers are proposed to memorize long
sentences and better capture semantic meanings for language.

\textbf{Multi-modal Fusion.}
Visual and linguistic features lie in different domains, and require
additional operations to fuse into a joint representation.
The simplest fusion operations are concatenation, element-wise addition, and
multiplication.
While linguistic features represent the entire sentence, the set of visual
features only implicitly does so.  Additional operations on the visual
feature set are needed to obtain a single image-wide feature representation.
Pooling operations such as max or average pooling, or simple addition
achieve this outcome.
However, a potential issue is that not all local features evenly contribute
to the final prediction.  Some visual features are related to irrelevant
pixels or regions, which should be ignored or have less influence on the
pooled output.
Importantly, linguistic features extracted from human instructions can
provide the guidance in identifying latent regions whose information should
be preserved.  
Attention \cite{vaswani2017attention} in the form of self-attention
and cross-attention modules are widely used to build correlations for
features in the same domain and across different domains, respectively.
%

\subsection{Pretraining on Scene Graph Parsing}
The vision-and-language task learning framework considers mutli-modal
information to play different roles.
Vision captures the information of objects and their interactions, which
reflects potential robotic tasks in the scene.
Language provides context. It helps to narrow down or determine the target
task over the task space inferred from vision.
We apply this insight and propose to pretrain the visual encoder on scene
graph parsing to help learn generic features that encode attributes and
relationships for objects.
A scene graph $G$ consists of:
\begin{list}{-}{\leftmargin=1em \itemindent=0em}
  \item a set of bounding boxes $B = \{b_1, \cdots, b_k\}, b_i \in \mathbb{R}^{4}$;
  \item a set of corresponding attributes $A = \{a_1, \cdots, a_k\}$ where
  the tuple $a_i$ include object category $c_i$, affordance $f_i$ 
  and general attribute $t_i$; and
  \item a set of relationships $R = \{r_1, \cdots, r_j\}$ between bounding
  boxes.
\end{list}
Employing the Stacked Motif Network \cite{zellers2018neural}, we factorize the
probability of constructing the graph $G$ given the RGB image $I$ as 
\begin{equation} \label{sg}
  P(G\,|\,I) = P(B\,|\,I) P(A\,|\,B,I) P(R\,|\,A,B,I)
\end{equation}
The bounding box generation model $P(B\,|\,I)$ is based on the Faster R-CNN
object detection model \cite{ren2015faster}.
It is pre-trained on the proposed dataset as described in
Section\ref{sec:sgp_dataset} and keeps parameters frozen 
during the training stage for attribute and relation prediction.
The attribute prediction model $P(A\,|\,B,I)$ involves encoding contextual
representation for each bounding box and decoding corresponding attribute
information. 
Predicted bounding boxes $B$ will be ordered from left to right
by the central x-coordinate in the image and fed into an
biLSTM for learning contextual representation $C = \{c_1, \cdots, c_k\}$.
Another LSTM is employed to decode object category, affordance and attribute.
The relation prediction model $P(R\,|\,A,B,I)$ follows a similar design
and predicts relationships between each pair of bounding boxes. 
Implementation details and corresponding code are publicly provided in 
\cite{ivagit_vltl}(Scene Graph Parsing).

\subsection{Pretraining on Semantic Textual Similarity}
We propose to admit explicit human instructions and implicit human intents,
the latter which might require incorporating environmental information
for full understanding.
Explicit human instructions are divided into complete and incomplete
instructions.
The complete instruction describes ordered sub-steps with 
full actions and objects, while the incomplete one
has partial information. There are four main reasons for missing
information: missing object, missing action, high-level verb and 
anaphoric reference.
Though composed with different low-level words, explicit instruction and
implicit intent have the same high-dimensional semantic meaning in the
robotic task domain. 
Semantic textual similarity tackles determining how similar two texts' 
semantic meanings are.
We apply this insight and propose to pretrain the linguistic encoder
on semantic textual similarity between explicit human instructions 
and implicit human intents.

The Siamese network is a network consists of twin networks which 
take different inputs but are coupled by a common objective
function.
Following the design of Sentence-BERT \cite{reimers-2019-sentence-bert}, 
we employ BERT followed by a pooling layer as the language modeling network to
learn separate embeddings for each sentence in the pair.
With two embeddings, we compute cosine similarity between them and 
use the mean squared error as the objective function:
\begin{equation} \label{sts}
  \loss[sts](s_{ex}, s_{im}; \epsilon) = 
  \frac{1}{n} \sum_{i=1}^{n} (\frac{s_{ex} \cdot s_{im}}{\text{max}(\left\| s_{ex} \right\|_2 \cdot \left\| s_{im} \right\|_2, \epsilon)})^2
\end{equation}
where $s_{ex}$ and $s_{im}$ are embeddings for explicit instruction and
implicit intent, $n$ is the number of sentence pairs and 
$\epsilon$ is set to $1e-8$.
The implementation is open-source \cite{ivagit_vltl}(Semantic Textual
Similarity).

\subsection{Instruction Following Framework}
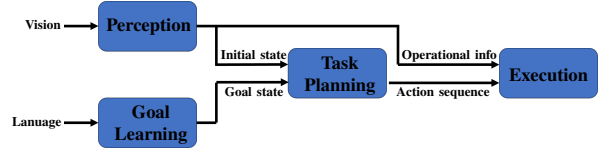
\begin{figure}[t]
  \centering
  \begin{tikzpicture} [outer sep=0pt, inner sep=0pt]
  \node[anchor=north west] (a) at (0in,0in) 
    {\includegraphics[width=0.8\columnwidth, clip=true,trim=0in 0in 0in 0in]{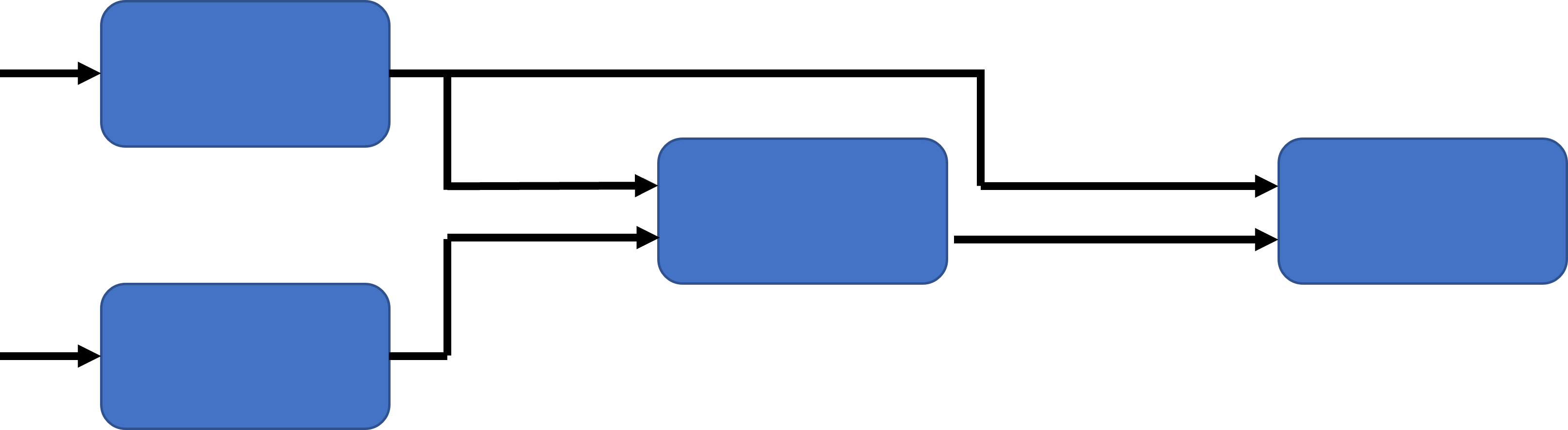}}; 
	\node[anchor=north west,xshift=-18pt,yshift=-4.pt,text=black,inner sep=3pt] at (a.north west) {\tiny \bf Vision};
	\node[anchor=north west,xshift=-23pt,yshift=-41.pt,text=black,inner sep=3pt] at (a.north west) {\tiny \bf Lanuage};
	
	\node[anchor=north west,xshift=12.7pt,yshift=-3.5pt,text=black,inner sep=3pt] at (a.north west) {\scriptsize \bf Perception};
	\node[anchor=north west,xshift=11.5pt,yshift=-36.5pt,text=black,inner sep=3pt] at (a.north west) {\parbox{0.5in}{\scriptsize \bf
    \centering Goal Learning}};
	\node[anchor=north west,xshift=83.pt,yshift=-18.1pt,text=black,inner sep=3pt] at (a.north west) {\parbox{0.5in}{\scriptsize \bf
    \centering Task Planning}};
	\node[anchor=north west,xshift=165.pt,yshift=-22.5pt,text=black,inner sep=3pt] at (a.north west) {\scriptsize \bf Execution};
	
	\node[anchor=north west,xshift=56.pt,yshift=-15.5pt,text=black,inner sep=3pt] at (a.north west) {\tiny \bf Initial state};
	\node[anchor=north west,xshift=57.5pt,yshift=-29.5pt,text=black,inner sep=3pt] at (a.north west) {\tiny \bf Goal state};
	
	\node[anchor=north west,xshift=124.5pt,yshift=-15.5pt,text=black,inner sep=3pt] at (a.north west) {\tiny \bf Operational info};
	\node[anchor=north west,xshift=122.5pt,yshift=-29.5pt,text=black,inner sep=3pt] at (a.north west) {\tiny \bf Action sequence};

  \end{tikzpicture}
  \caption{Human instruction following framework. This robotic framework 
  is designed for performing manipulation tasks by following instructions 
  from human. It takes vision and language as input and performs a sequence
  of actions in the real world.
  \label{fig_il}}
\end{figure}

As shown in the Figure \ref{fig_il}, the proposed hybrid, modular 
instruction following framework consists of four components: 
perception, goal learning, task planning and execution.
The hybrid design leverages the strength of semantic feature learning
from deep neural networks and the strength of symbolic manipulation
from symbolic planners, which compensates limitation of both methods. 
Modular design benefits include: 
easy analysis of which part leads to failure; 
easy component replacement with better methods; and 
easy augmentation with other components, such as life-long learning; to make
the entire framework more complete and powerful.
The \textbf{Perception} module is responsible for interpreting 
visual information of surrounding environments. 
The \textbf{Goal Learning} module learns symbolic goal representation
for the \textbf{Task Planning} module.
The \textbf{Task Planning} module generates a sequence of low-level actions.
The \textbf{Execution} module performs generated actions with operational
information detected by the \textbf{Perception} module.

This work uses Mask R-CNN \cite{he2017mask} as the \textbf{Perception}
module to detect objects and their category segmentation masks.
The categorical information is detected and 
corresponding affordances and attributes are retrieved from knowledge base
to build the initial state for PDDL.
A vision-and-language learning network is the \textbf{Goal Learning} module
for goal state prediction.
The \textbf{Task Planner} module is a PDDL planner that outputs
a primitive action sequence from the detected initial and goal states,
which is then sent to the robot for execution.
Robotic action requires operational information, which is provided via
masks detected by \textbf{Perception} module.

\section{Vision-and-Language Model Benchmark}
This section introduces three training datasets with corresponding training
policies for learning three tasks.
The evaluation metric is then discussed along with benchmarking to
evaluate each component in the vision-and-language goal learning framework,
and the two proposed pretraining tasks.

\subsection{Datasets}
The proposed three datasets are symbolic goal learning,
scene graph parsing and semantic textual similarity datasets.  
The dataset and training code is provided \cite{ivagit_vltl} (Dataset).

\subsubsection{Symbolic Goal Learning Dataset}
For learning symbolic goal representation from vision and language, 
we created a dataset containing 32,070 images paired with natural language,
which could be either an explicit instruction or implicit intent.
It covers five daily activities: picking and placing, object delivery,
cutting, cooking, and cleaning.
We employ the simulator AI2THOR to generate image and
sentence pairs, which is automatically annotated with PDDL goal states.
Besides imperfect natural lanugage, we also include imperfect vision where
one or both objects involved in the task not exist in the image.
With such input, the vision-and-language symbolic goal learning network is
expected to predict the missing object to be ``unknown'' in the goal state
output.

\subsubsection{Scene Graph Parsing Dataset \label{sec:sgp_dataset}}
The Scene Graph Parsing dataset was also created using AI2THOR and focuses
on common daily objects.  Object categories, affordances, attributes and
their relationships are recorded.
It covers 32 categories, 4 affordances, 5 attributes and 4 relationships. 
The total dataset includes 32,070 RGB images, 
automatically annotated with ground-truth bounding box, category,
affordance, attribute and relationship, which is labor-free.

\subsubsection{Semantic Textual Similarity Dataset \label{sec:sts_dataset}}
The created Semantic Textual Similarity dataset is for learning the similar
semantic meaning between explicit human instructions and implicit human
intents for robotic tasks.
There same five daily activities are in the dataset: picking and placing,
object delivery, cutting, cleaning, and cooking. 
It contains 90,000 pairs of explicit instruction and implicit intent,
generated from a manually created list of templates.
For the purpose of improving the diversity, sentences are automatically
paraphrased by Parrot \cite{prithivida2021parrot} during the generation
process.
To automatically rank the similarity of two sentences, three different
scores are assigned based on the following rules:
\begin{itemize}
  \item 5.0 if two sentences contain the same subject and object,
  \item 3.3 if two sentences match either subject or object, and
  \item 1.7 if two sentences describe the same task.
\end{itemize}

\subsection{Training Policy}
All models evaluated in Table \tabref{benchmark} are implemented in Pytorch
and trained on a single nVidia Titax-X (Pascal).
For training, Adamax and AdamW are the optimizers for the LSTM and
BERT models with 1e-8 and 5e-5 for initial learning rate. 
The learning rate scheduler is warmup cosine with a 1 epoch warmup
stage.  The model is trained for 20 epoches.
For model implementations and training details, see \cite{ivagit_vltl}.

\subsection{Evaluation Metric}
Evaluating the prediction accuracy of the vision-and-language models
should test for symbolic matching to the PDDL goal state entities, which
are action, subject, and object.  We propose the Robotic Goal Learning
(RGL) accuracy score:
\begin{equation} \label{rgl}
  \text{RGL} = \delta(\hat{a}, a, \hat{s}, s, \hat{o}, o) \equiv
  \delta(\hat{a},a)\cdot\delta(\hat{s},s)\cdot\delta(\hat{o},o).
\end{equation}
where $\delta(\cdot)$ is the Kronecker delta function, $\hat{a}$ and $a$ are predicted and ground-truth action label,
and the same holds for the subject $s$ and object $o$ labels.

\begin{table}[t]
  \centering
  \caption {Benchmarking Vision-and-Language Symbolic Goal Learning\tablabel{benchmark}}
  \small
  \setlength{\tabcolsep}{2.5pt}
  \begin{tabular}{ | l | c | c |}
    \hline
    \bf{Model}              & \bf{RGL Accuracy(\%)}     & Speed (fps) \\ \hline 
Grid-LSTM-Concat            &  77.24         &     476.19    \\ \hline
Grid-LSTM-Add               &  80.29         &     476.19    \\ \hline
Grid-LSTM-Mul               &  72.47         &     476.19    \\ \hline
Region-LSTM-Concat          &  81.35         &       9.14    \\ \hline
Grid-BERT-Concat            &  89.02         &     230.95    \\ \hline
Grid-BERT-Add               &  85.69         &     226.41    \\ \hline
\hline
Grid-BERT-TDAtt-v1\cite{anderson2018bottom}  &  92.61   &  215.98    \\ \hline
Grid-BERT-TDAtt-v2\cite{anderson2018bottom}  &  93.54   &  210.71    \\ \hline
Grid-BERT-CoAtt\cite{yu2019deep}             &  92.30   &  120.70 \\ \hline \hline
Grid-BERT-Concat (SGP)      &  93.55         &     230.95    \\ \hline
Grid-BERT-Concat (STS)      &  90.96         &     230.95    \\ \hline
Grid-BERT-Concat (SGP\&STS) &  94.54         &     230.95    \\ \hline
  \end{tabular}
\end{table}

\subsection{Benchmarking Vision-and-Language Goal Learning}

The test configurations in Table \tabref{benchmark} permit comparison of
different implementation choices regarding the core components, plus the
effect of attention models and pre-training. The baseline visual and 
and language encoders will employ Grid features and LSTM, respectively. 
For reference, the LSTM-only and BERT-only models perform at 67.07\% and
55.91\%. 

Regarding the fusion component for the baseline model, three simple
strategies were tested: concatenation (concat), addition (add) and
multiplication (mul).  The best of the three tested for the baseline
LSTM implementation is addition.  Switching from Grid to Region
features, with concatenation, leads to a small boost in performance of
4.11\% but a 50x drop in processing rate.
Grid feature encoders show better trade-off between prediction accuracy
and inference speed if real-time is important.
Considering a change in the language encoder to BERT, there is a boost
in performance to 89.02\% and 85.69\%, for fusion by concatenation and
addition, respectively. The 2x drop in timing is not serious, thus
BERT+concat would be the more sensible option to use. 
It provides a 11.78\% boost in performance and still operates beyond
frame-rate.

Regarding attention versus pre-training, the attention model implemented
were Top-Down Attention (TDAtt)\cite{anderson2018bottom} and 
Co-Attention (CoAtt)\cite{yu2019deep}.  There two top-down attention
variants: directly feeding the fused embedding for classification, 
and concatenating the fused embedding with extracted visual and linguistic
features.
Pre-training involved Scene Graph Parsing (SGP) and Semantic Textual
Similarity (STS) tasks, as noted earlier.
Of the two strategies for increasing performance, independent
pre-training of the two encoders provided the best boost without
affecting processing time.
While attention models did improve the outcomes, they are known to
require customized training policies to operate well
\cite{liu-etal-2020-understanding}.

\section{Manipulation Experiments}

\begin {table*}[t]
  \centering
  \caption {Results of manipulation experiments in AI2THOR.
  P: perception; GL: goal learning; TP: task planning; E: execution.
  \tablabel{me}}
  \small
  \setlength{\tabcolsep}{4.5pt}
  \resizebox{\textwidth}{!}{
  \begin{tabular}{ | c | c | c | c | c | c | c | c | c | c | c | c | c | c | c | c | c | c | c | c | c | c | c | c | c |}
    \hline
      & \multicolumn{4}{c|}{\bf Pick\_n\_Place}  & \multicolumn{4}{c|}{\bf Object Delivery}  & \multicolumn{4}{c|}{\bf Cut}  & \multicolumn{4}{c|}{\bf Cook}  & \multicolumn{4}{c|}{\bf Clean} &  \multicolumn{4}{c|}{\bf VSR (\%)} \\ \hline
      & P & GL & TP & E & P & GL & TP & E & P & GL & TP & E & P & GL & TP & E & P & GL & TP & E & P & GL & TP & E \\ \hline
 {\bf Easy} & 9 & 9 & 9 & 9 & 10 & 9 & 9 & 9 & 10 & 10 & 10 & 10 & 10 & 8 & 8 & 8 & 10 & 10 & 10 & 10 & 98.0 & 92.0 & 92.0 & 92.0 \\ \hline    
 {\bf Medium} & 10 & 8 & 8 & 8 & 9 & 9 & 8 & 8 & 10 & 8 & 8 & 8 & 10 & 7 & 7 & 7 & 10 & 10 & 10 & 9 & 98.0 & 84.0 & 82.0 & 80.0 \\ \hline    
 {\bf Hard1} & 10 & 7 & 7 & 7 & 10 & 8 & 8 & 8 & 10 & 8 & 8 & 8 & 9 & 7 & 6 & 6 & 9 & 7 & 7 & 6 & 96.0 & 74.0 & 72.0 & 70.0\\ \hline    
 {\bf VSR (\%)} & 96.7  & 80.0 & 80.0 & 80.0 & 96.7 & 86.7 & 83.3 & 83.3 & 100.0 & 86.7 & 86.7 & 86.7 & 96.7 & 73.3 & 70.0 & 70.0 & 96.7 & 90.0 & 90.0 & 83.3 & 97.3 & 83.3 & 82.0 & 80.7 \\ \hline \hline
 \multicolumn{21}{|c|}{} & \multicolumn{4}{c|}{\bf ISR (\%)} \\ \hline
 {\bf Hard2} & 9 & 7 & 10 & 10 & 10 & 6 & 10 & 10 & 10 & 8 & 10 & 10 & 9 & 5 & 10 & 10 & 10 & 9 & 10 & 10 & 96.0 & 70.0 & 100.0 & 100.0 \\ \hline \hline
 \multicolumn{21}{|c|}{} & \multicolumn{4}{c|}{\bf SR (\%)} \\ \hline
 {\bf SR (\%)} & 95.0 & 77.5 & 85.0 & 80.0 & 97.5 & 80.0 & 87.5 & 83.3 & 100.0 & 85.0 & 90.0 & 86.7 & 95.0 & 67.5 & 77.5 & 70.0 & 97.5 & 90.0 & 92.5 & 83.3 & 97.0 & 80.0 & 86.5 & 85.5\\ \hline        
  \end{tabular}}
  \vspace*{-1.25em}
\end {table*}

\begin {table}
  \caption {Comparison of manipulation experiments to existing methods
  \tablabel{cme}}
  \small
  \setlength{\tabcolsep}{4.5pt}
  \resizebox{\columnwidth}{!}{
  \begin{tabular}{ | c | c | c | c | c | c | c | c | c | c | c | c | }
    \hline
      & \multicolumn{2}{c|}{V2A \cite{nazarczuk2020v2a}}  & \multicolumn{2}{c|}{ALFRED \cite{shridhar2020alfred}}  & \multicolumn{2}{c|}{Mod. \cite{corona2020modular}}  & \multicolumn{2}{c|}{HiTUT \cite{zhang2021hierarchical}} & \multicolumn{3}{c|}{SGL} \\ \hline
      & S & U & S & U & S & U & S & U & S\_GL & U\_GL & U\_E \\ \hline
 {\bf SR (\%)} & 44.7  &  40.2 & 70.3  & 49.9  & 71.9  & 63.0  & 87.7  & 80.6  & 94.5  & 80.0 & 85.5 \\ \hline
 {\bf VSR (\%)} & 23.8  & 6.4  & 70.3  & 49.9  & 71.9  & 63.0  & 87.7  & 80.6  & 96.7 & 83.3 & 80.7 \\ \hline
 {\bf ISR (\%)} & 65.0  &  79.9 & -  & -  & -  & -  & -  & -  & 88.8 & 70.0 & 100.0 \\ \hline       
  \end{tabular}}
  \vspace*{-1.25em}
\end {table}

\subsection{Experimental Setup}
Manipulation experiments in AI2THOR evaluate the robustness and 
generalization of the proposed instruction following framework to 
novel scenarios.
Five different daily activities are conducted, which include
Picking and Placing, Object Delivery, Cutting, 
Cleaning and Cooking.
There are four different levels of scenarios for each task.
Easy scenario only contains involved objects in the scene.
Medium scenario incorporates irrelevant objects.
The first hard scenario further includes multiple candidates while 
the second hard scenario misses partial or all objects required to perform 
the task.
Due to missing objects in the scene, task planning is not expected to 
find valid solutions and execution is also not required for the
second hard case.
There are 10 scenarios for each level and either novel instruction or 
intent will be paired with the image.
The model, which consists of grid feature encoder, BERT and concatenation 
and is pretrained on both tasks, are employed.

\subsection{Manipulation Metrics}
To evaluate each module in the instruction following framework, 
each manipulation experiment trial is considered as successful 
if it satifies four conditions.
For \textbf{Perception}, all involved objects are required to be
correctly detected, which constructs the initial state for PDDL.
For \textbf{Goal Learning}, PDDL goal state should be correctly predicted.
For \textbf{Task Planning}, generated action sequence is composed of
correct ordered actions.
Given that AI2THOR does not support physical modeling of robot-object
interaction, \textbf{Execution} evaluation requires the
Intersection-of-Union (IoU) of detected and ground-truth masks for objects
to be over the 0.5 threshold. 
Based on \cite{nazarczuk2020v2a}, Valid Success Rate
(VSR) and Invalid Success Rate (ISR) are employed for easy, medium and the first hard, and the second hard scenarios, respectively. 
VSR evaluates tasks with valid solutions while ISR evaluates ones where
there is no valid solution.
Success Rate (SR) is used to take the average over all valid and invalid
tasks. 
%
%

\subsection{Outcomes and Analysis for Manipulation Experiments}
Results of manipulation experiments are collected in the Table \tabref{me}. 
Evaluating the performance change between seen and unseen scenarios for 
the proposed symbolic goal learning network, success rate drops from 
94.5\% to 80.0\%. 
The performance drop is mainly caused by two aspects. 
Firstly, training dataset doesn't include images with multiple 
candidate objects, which causes the domain shift. 
The grid-based feature encoder may have trouble localizing the correct
regions of interest. 
Secondly, the main issues is with the \textit{cook} tasks. 
The potential reason could be the appearance of the microwave and stove
burner in the same image leading to a misunderstanding of cookware.

For easy, medium and the first hard scenarios which have valid solutions 
for tasks, results show that success rate of task planning is roughly 
equal to the product of perception and goal learning. 
The observation shows the approximate independence between 
perception and goal learning modules. 
The 97.0\% success rate for perception and 1\% performance drop from
task planning to execution show that the existing perception module
works pretty well. 
The goal learning module, which is the primary interest here, is the
biggest bottleneck in the current stage. 
Further study of symbolic goal learning framework could provide the
proposed instruction following framework with significant boost.

For the second hard scenario which has no valid solutions or plans, 
the result shows that success rate of task planning is higher than the
product of perception and goal learning. 
The reason is that even when goal learning module fails to predict a
missing object as \textit{unknown}, the perception module does not
detect the missing object. 
With an incomplete initial representation the symbolic task planner 
correctly outputs \textit{no solution}, which shows the value of the
modular system and of symbolic planning. 
With a symbolic module computing the sequence of primitive actions, the
system knows whether it is possible to achieve the task instead of
having to predict the sequential actions via connectionist approaches.

\subsection{Outcomes and Analysis for Comparison}
Since the proposed method focuses on manipulation tasks which 
do not include navigation, we collect experimental results of 
manipulation sub-tasks for existing connectionist approaches 
in Table \tabref{cme} for perform approximate comparative analysis. 
Seen and Unseen scenarios are denoted as S and U.
As shown in the Table \tabref{cme}, the proposed framework 
achieves average 85.5\% task success rate which 
outperforms all methods, thereby supporting the hypothesized benefits
of the proposed hybrid, modular instruction following framework.

For seen to unseen goal learning (GL), the proposed method experiences
average 14.5\% performance drop, 13.4\% performance drop for tasks with
valid solution and 18.8\% performdance drop for task with no valid
solutions. 
The result shows that the proposed symbolic goal learning 
achieves better performance on valid cases than invalid case. 
The invalid case requires the model to first interpret visual
information and then use it to correct mismatched language, which is
challenging.  However the task planner fully compensates.
SGL achieves a lower performance drop in success rate than ALFRED and a
lower performance drop in VSR than V2A. 
The result shows that symbolic goal learning via deep neural networks
experiences a lower performance drop from seen to unseen than predicting
sequential actions. 
The performance drop of Mod. \cite{nazarczuk2020v2a} and 
HiTUT \cite{zhang2021hierarchical} is 5.6\% and 7.4\% 
lower than the proposed method, SGL. 
Mod. segments the entire instruction into several sub-tasks, which helps
significantly improve its performance on unseen scenarios and suggests
that SGL should pre-process long-length instructions into task-specific
segments.
HiTUT incorporates self-monitoring and backtracking which allows the
robotic agent to perform the action again if it fails the previous
attempt. 
The success of self-monitoring and backtracking indicates that
incorporating similar modules within SGL to deal with dynamic
environments may improve performance.

\section{Conclusion}

To address human instruction following with diverse natural language
inputs, we propose to compensate for implicit or missing information via
vision and present a hybrid, modular framework consisting of symbolic goal
learning via deep netural networks and task planning via symbolic planners.
We propose a vision-and-language goal learning framework, which consists of
the visual encoder, linguistic encoder, multi-modal fusion and
classification.
Benchmarking compares the impacts of different techniques for the
different components.
For learning generic features and boosting the performance when fine-tuning
on specific tasks, we propose to separately pretrain the visual and 
linguistic encoder on scene graph parsing and semantic textual similarity
tasks.
We show the effectiveness of the two pretraining tasks on a model with 
visual grid features, BERT, and fusion by concatenation,
Evaluation of the instruction following framework in the AI2THOR simulator
shows robustness to novel scenarios. 
The hybrid framework combines the strength of semantic feature learning 
from deep neural networks and capability of rejecting invalid tasks
from symbolic planners.
The modular design of the framework enables easy determination and analysis 
of the cause of failure, simple replacement of each component, and 
incorporation of more modules.
The current proposed framework lacks modules such as a task progress monitor
and feedback mechanism to deal with dynamic environments, which will be the
aim of future work.
Additionally, the proposed symbolic goal learning network is trained on
synthetic data; we will work on domain adaptation to bridge the gap between
synthetic and real data.

\bibliographystyle{IEEEtran}
\bibliography{regular,crowdsourcing}
\end{document}